\documentclass[11pt,a4paper]{article}
\usepackage[hyperref]{emnlp2020}
\usepackage{times}
\usepackage{latexsym}

\usepackage{microtype}

\usepackage{graphicx}
\usepackage{epstopdf}
\usepackage{amsmath}
\usepackage{amssymb}
\usepackage{multirow}
\usepackage{booktabs}

\usepackage{lipsum}

\aclfinalcopy % Uncomment this line for the final submission
\title{CSP: Code-Switching Pre-training for Neural Machine Translation}

\author{Zhen Yang \\
  Affiliation / Address line 1 \\
  Affiliation / Address line 2 \\
  Affiliation / Address line 3 \\
  \texttt{email@domain} \\\And
  Second Author \\
  Affiliation / Address line 1 \\
  Affiliation / Address line 2 \\
  Affiliation / Address line 3 \\
  \texttt{email@domain} \\}

\author{Zhen Yang,  Bojie Hu, Ambyera Han, Shen Huang and Qi Ju\footnotemark[1] \\
  Tencent Minority-Mandarin Translation \\
  {\tt \{zieenyang, bojiehu, ambyera, springhuang, damonju\}@tencent.com}}
  
\date{}

\begin{document}
\maketitle
\newcommand\blfootnote[1]{%
\begingroup
\renewcommand\thefootnote{}\footnote{#1}%
\addtocounter{footnote}{-1}%
\endgroup
}
\blfootnote{* indicates corresponding author.}

\begin{abstract}
 This paper proposes a new pre-training method, called Code-Switching Pre-training (CSP for short) for Neural Machine Translation (NMT). Unlike traditional pre-training method which randomly masks some fragments of the input sentence,  the proposed CSP randomly replaces some words in the source sentence with their translation words in the target language. Specifically, we firstly perform lexicon induction with unsupervised word embedding mapping between the source and target languages, and then randomly replace some words in the input sentence with their translation words according to the extracted translation lexicons. CSP adopts the encoder-decoder framework: its encoder takes the code-mixed sentence as input, and its decoder predicts the replaced fragment of the input sentence. In this way, CSP is able to pre-train the NMT model by explicitly making the most of the cross-lingual alignment information extracted from the source and target monolingual corpus. Additionally,  we relieve the pretrain-finetune discrepancy caused by the artificial symbols like [mask].  To verify the effectiveness of the proposed method, we conduct extensive experiments on unsupervised and supervised NMT. Experimental results show that CSP achieves significant improvements over baselines without pre-training or with other pre-training methods.
\end{abstract}

\section{Introduction}
Neural machine translation \cite{Kalchbrenner:13,sutskever:14,cho:14b,bahdanau:14} which typically follows the encoder-decoder framework, directly applies a single neural network to transform the source sentence into the target sentence. With tens of millions of trainable parameters in the NMT model, translation tasks are usually data-hungry, and many of them are low-resource or even zero-resource in terms of training data. Following the idea of unsupervised and self-supervised pre-training methods in the NLP area \cite{peters-etal-2018-deep,radford2018improving,radford2019language,devlin2018bert,yang2019xlnet}, some works are proposed to improve the NMT model with pre-training, by making full use of the widely available monolingual corpora \cite{lample2019cross,song2019mass,edunov2019pre,huang-etal-2019-glossbert,wang-etal-2019-denoising,rothe2019leveraging,clinchant2019use}. Typically, two different branches of pre-training approaches are proposed for NMT: \textbf{\emph{model-fusion}} and \textbf{\emph{parameter-initialization}}.

The \textbf{\emph{model-fusion}} approaches seek to incorporate the sentence representation provided by the pre-trained model, such as BERT, into the NMT model \cite{yang2019towards,clinchant2019use,weng2019improving,zhu2020incorporating,lewis2019bart,liu2020multilingual}.  These approaches are able to leverage the publicly available pre-trained checkpoints in the website but they need to change the NMT model to fuse the sentence embedding calculated by the pre-trained model. Large-scale parameters of the pre-trained model significantly increase the storage cost and inference time, which makes it hard for this branch of approaches to be directly used in production.  As opposed to \textbf{\emph{model-fusion}} approaches, the \textbf{\emph{parameter-initialization}} approaches aim to directly pre-train the whole or part of the NMT model with tailored objectives, and then initialize the NMT model with pre-trained parameters \cite{lample2019cross,song2019mass}. These approaches are more production-ready since they keep the size and structure of the model same as standard NMT systems.

While achieving substantial improvements, these pre-training approaches have two main cons. Firstly, as pointed out by \citet{yang2019xlnet}, the artificial symbols like [mask] used by these approaches during pre-training are absent from real data at fine-tuning time, resulting in a pretrain-finetune discrepancy. Secondly, while each pre-training step only involves sentences from the same language, these approaches are unable to make use of the cross-lingual alignment information contained in the source and target monolingual corpus. We argue that, as a cross-lingual sequence generation task, NMT requires a tailored pre-training objective which is capable of making use of cross-lingual alignment signals explicitly, e.g., word-pair information extracted from the source and target monolingual corpus, to improve the performance.

To address the limitations mentioned above, we propose Code-Switching Pre-training (CSP) for NMT. We extract the word-pair alignment information from the source and target monolingual corpus automatically, and then apply the extracted alignment information to enhance the pre-training performance. The detailed training process of CSP can be presented in two steps:  1) perform lexicon induction to get translation lexicons by unsupervised word embedding mapping \cite{artetxe2018acl,ConneauWord}; 2) randomly replace some words in the input sentence with their translation words in the extracted translation lexicons and train the NMT model to predict the replaced words. CSP adopts the encoder-decoder framework: its encoder takes the code-mixed sentence as input, and its decoder predicts the replaced fragments based on the context calculated by the encoder. By predicting the sentence fragment which is replaced on the encoder side, CSP is able to either attend to the remaining words in the source language or to the translation words of the replaced fragment in the target language. Therefore, CSP trains the NMT model to: 1) learn how to build the sentence representation for the input sentence as the traditional pre-training methods do; 2) learn how to perform cross-lingual translation with extracted word-pair alignment information. In summary, we mainly make the following contributions:
\begin{itemize}
\item We propose the code-switching pre-training for NMT, which makes full use of the cross-lingual alignment information contained in source and target monolingual corpus to improve the pre-training for NMT. 
\item We conduct extensive experiments on supervised and unsupervised translation tasks. Experimental results show that the proposed approach consistently achieves substantial improvements.
\item Last but not least, we find that CSP can successfully handle the code-switching inputs. 
\end{itemize}

\footnotetext[1]{To be used in production easily, these models need to be distilled into a student model with the structure and size same as standard NMT systems.}
\section{Related works}
Several approaches have been proposed to improve NMT with pre-training. \citet{edunov2019pre} proposed to feed the last layer of ELMo to the encoder of NMT and investigated several different ways to add pre-trained language model representations to the NMT model. \citet{weng2019improving} proposed a bi-directional self-attention language model to get sentence representation and introduced two individual methods, namely weighted-fusion mechanism and knowledge transfer paradigm,  to enhance the encoder and decoder.  \citet{yang2019towards} proposed a concerted training framework to make the most use of BERT in NMT. \citet{zhu2020incorporating} proposed to fuse the representations from BERT with
each layer of the encoder and decoder of the NMT model through attention mechanisms. Large-scale parameters of the pre-trained model in these approaches discussed above significantly increase the storage cost and inference time, which makes these approaches a little far from production.\footnotemark[1] The other branch of approaches aims to keep the structure and size the same to the standard NMT system and designs some pre-training objectives tailored for NMT. \citet{lample2019cross} proposed Cross-Lingual Language Model (XLM) objective and built a universal cross-lingual encoder. To improve the cross-lingual pre-training, they introduced supervised translation language modeling objective relying on the parallel data available. \citet{song2019mass} proposed the MASS objective to pre-train the whole NMT model instead of only pre-training the encoder by XLM. CSP builds on top of \citet{lample2019cross} and \citet{song2019mass}, and it explicitly makes full use of the alignment information extracted from the source and target monolingual corpus to enhance pre-training.

There have also been works on applying pre-specified translation lexicons to improve the performance of NMT.  \citet{hokamp2017lexically} and \citet{post2018fast} proposed an altered beam search algorithm, which took target-side pre-specified translations as lexical constraints during beam search.  \citet{song2019code} investigated a data augmentation method, making code-switched training data by replacing source phrases with their target translations according to the pre-specified translation lexicons. Recently, motivated by the success of unsupervised cross-lingual embeddings, \citet{ArtetxeUnsupervised}, \citet{Lample2017Unsupervised} and \citet{yang-etal-2018-unsupervised} applied the pre-trained translation lexicons to initialize the word embeddings of the unsupervised NMT model. \citet{haipeng2019acl} applied translation lexicons to unsupervised domain adaptation in NMT. In this paper, we apply the translation lexicons automatically extracted from the monolingual corpus to improve the pre-training of NMT.

\begin{figure*}[ht]
   \begin{center}
   \includegraphics[scale=0.54]{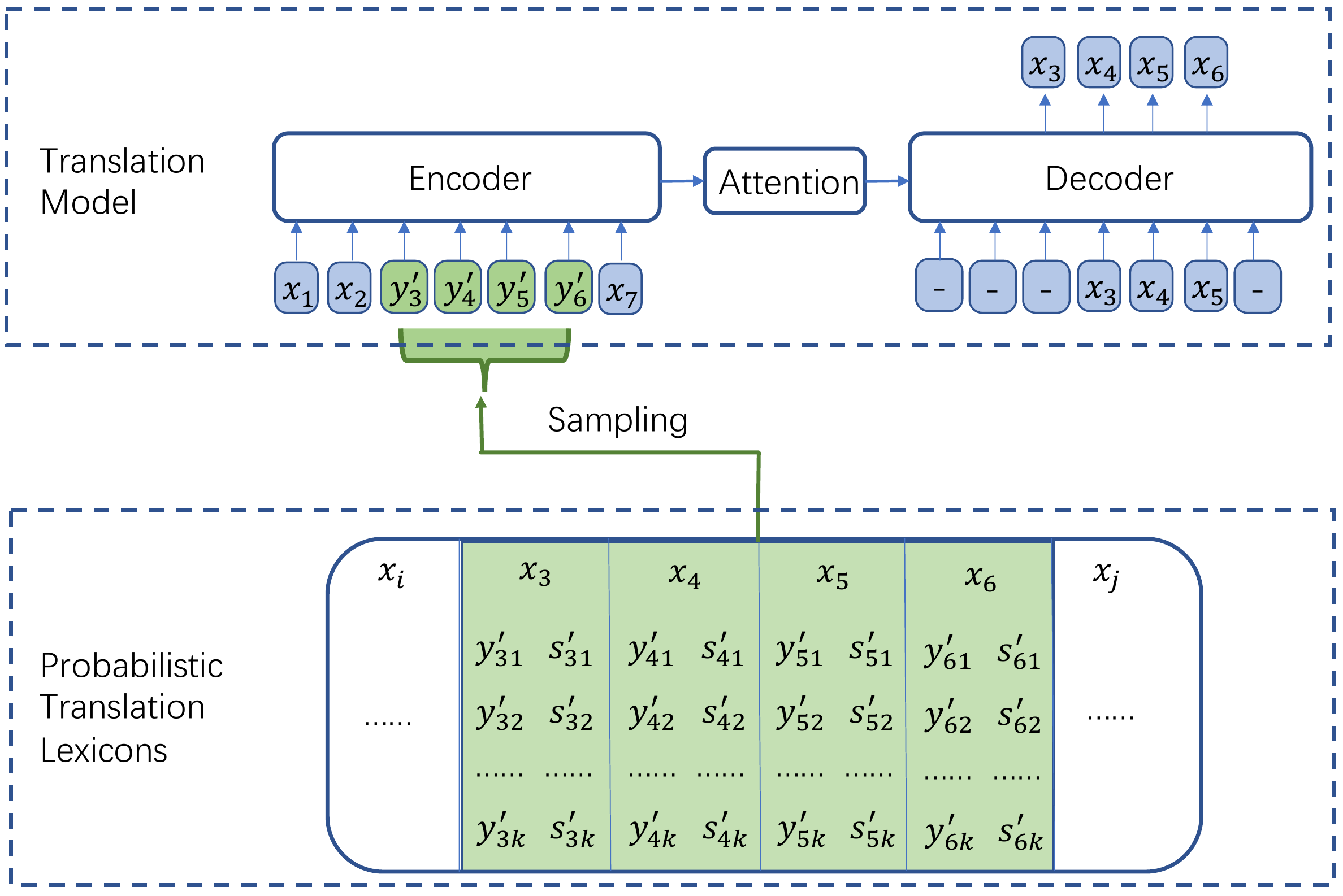}
   \end{center}
   \caption{\label{fig:CMP_EXA} The training example of our proposed CSP which randomly replaces some words in the source input with their translation words based on the probabilistic translation lexicons. Identical to MAS, the token ‘-’ represents the padding in the decoder. The attention module represents the attention between the encoder and decoder}	
\end{figure*}

\section{CSP}
In this section, we firstly describe how to build the shared vocabulary for the NMT model; then we present the way extracting the probabilistic translation lexicons; and we introduce the detailed training process of CSP finally. 

\subsection{Shared sub-word vocabulary}
This paper processes the source and target languages with the same shared vocabulary created through the sub-word toolkits, such as Sentence-Piece (SP) and Byte-Pair Encoding (BPE) \cite{sennrich-etal-2016-neural}. We learn the sub-word splits on the concatenation of the sentences equally sampled from the source and target corpus. The motivation behind is two-fold: Firstly, with processing the source and target languages by the shared vocabulary, the encoder of the NMT model is able to share the same vocabulary with the decoder. Sharing the vocabulary between the encoder and decoder makes it possible for CSP to replace the source words in the input sentence with their translation words in the target language. Secondly, as pointed out by \citet{lample2019cross}, the shared vocabulary greatly improves the alignment of embedding spaces. 

\subsection{Probabilistic translation lexicons}
\label{ptl}
 Recently, some works successfully learned translation equivalences between word pairs from two monolingual corpus and  extracted translation lexicons \cite{artetxe2018acl,ConneauWord}. Following \citet{artetxe2018acl}, we utilize unsupervised word embedding mapping to extract probabilistic translation lexicons with monolingual corpus only. The probabilistic translation lexicons in this paper are defined as one-to-many source-target word translations. Specifically, giving separate source and target word embeddings, i.e., $X_e$ and $Y_e$ trained on source and target monolingual corpus $X$ and $Y$, unsupervised word embedding mapping utilizes self-learning or adversarial-training to learn a mapping function $f(X)=WX$, which transforms source and target word embeddings to a shared embedding space. With word embeddings in the same latent space, we measure the similarities between source and target words with the cosine distance of word embeddings. Then, we extract the probabilistic translation lexicons by selecting the top $k$ nearest neighbors in the shared embedding space. Formally, considering the word $x_i$ in the source language,  its top $k$ nearest neighbor words in the target language, denoted as $y^{'}_{i1},y^{'}_{i2},\ldots,y^{'}_{ik}$ are extracted as its translation words, and the corresponding normalized  similarities $s^{'}_{i1},s^{'}_{i2},\ldots,s^{'}_{ik}$ are defined as the translation probabilities.  
 
\subsection{Training process of CSP}
CSP only requires monolingual data to pre-train the NMT model. Given an unpaired source sentence $x \in X$, where $x=(x_1,x_2,\ldots,x_m)$ is the source sentence with $m$ tokens, we denote $x_{[u:v]}$ as the sentence fragment of $x$ from $u$ to $v$ where $0<u<v<m$, and denote $x^{\backslash u:v}$ as modified version of $x$ where its fragment from position $u$ to $v$ are replaced with their translation words according to the probabilistic translation lexicons. Formally, $x^{\backslash u:v}$ is represented as:
\begin{equation}
x^{\backslash u:v}=(x_1,\ldots,x_{u-1},y^{'}_{u},\ldots,y^{'}_{v},x_{v+1}\ldots,x_m)
\end{equation}
where $x^{\backslash u:v}_{[u:v]}=(y^{'}_u,\ldots,y^{'}_v)$ is sampled based on the extracted probabilistic translation lexicons presented on Section \ref{ptl}. Here, we take the replacing process from $x_u$ to $y^{'}_u$ as an example. Considering the source word $x_u$, its top $k$ translation words $y^{'}_{u1},y^{'}_{u2},\ldots,y^{'}_{uk}$ and the translation probabilities $s^{'}_{u1},s^{'}_{u2},\ldots,s^{'}_{uk}$, $y^{'}_u$ is calculated as:
\begin{equation}
y^{'}_u = y^{'}_{uj}(1\leq j \leq k)
\end{equation}
where $y^{'}_{uj}$ is decided by performing multinomial sampling on the distribution defined by translation probabilities $s^{'}_{u1},s^{'}_{u2},\ldots,s^{'}_{uk}$. With higher translation probability $s^{'}_{uj}$, the translation word $y^{'}_{uj}$ is more likely to be selected.

Similar to \citet{song2019mass}, CSP pre-trains a sequence to sequence model by predicting the sentence fragment $x_{[u:v]}$ with the modified sequence $x^{\backslash u:v}$ as input. With the log likelihood as the objective function, CSP trains the NMT model on the monolingual corpora $X$ as:
\begin{equation}
\begin{array}{lcl}
L(\theta;X)= \frac{1}{|X|}\sum_{x \in X} log P(x_{[u:v]}|x^{\backslash u:v};\theta) &\\
=  \frac{1}{|X|}\sum_{x \in X} log \prod\limits_{t=u}^{v} P(x_t|x_{<t},x^{\backslash u:v};\theta)
\end{array}
\end{equation}
Figure \ref{fig:CMP_EXA} shows an example for CSP training, where the original source sentence $(x_1,x_2,x_3,x_4,x_5,x_6,x_7)$ with the fragment $(x_3,x_4,x_5,x_6)$ being replaced with their translation words, i.e., $(y^{'}_3,y^{'}_4,y^{'}_5,y^{'}_6)$ sampled from the extracted probabilistic translation lexicons. The encoder takes the code-mixed source sentence as input, and the decoder only predicts the replaced fragment $(x_3,x_4,x_5,x_6)$.  

\section{Experiments and Results}
This section describes the experimental details about CSP pre-training and fine-tuning on the supervised and unsupervised NMT tasks. To test the effectiveness and generality of CSP, we conduct extensive experiments on English-German, English-French and Chinese-to-English translation tasks.

\begin{table*}[htb]
			\centering
			  \scalebox{0.95}{
				\begin{tabular}{c|ccccc}
					\toprule[2pt]
					 System & en-de	& de-en	&	en-fr &   fr-en & zh-en\\
					\midrule[1pt]
					\citet{yang-etal-2018-unsupervised}  & 10.86& 14.62&16.97 & 15.58 & 14.52  \\
                    \citet{lample-etal-2018-phrase}  &17.16 & 21.0 & 25.14 & 24.18  & - \\
                    \midrule[1pt]
                    \citet{lample2019cross} & 27.0& 34.3&33.4 & 33.3&- \\
                    \citet{song2019mass} &28.1 & 35.0& 37.5& \textbf{34.6}& -\\
                    \midrule[1pt]
                    \citet{lample2019cross} (our reproduction)&27.3&33.8&32.9&33.5 & 22.1 \\
                    \citet{song2019mass} (our reproduction) & 27.9 & 34.7& 37.3 & 34.1& 22.8\\
                    \textbf{CSP and fine-tuning (ours)} &\textbf{28.7} & \textbf{35.7}& \textbf{37.9}& 34.5& \textbf{23.9} \\
					\bottomrule[2pt]
				\end{tabular}}
				\caption{\label{tab:Result_UNMT} The translation performance of the fine-tuned unsupervised NMT models. To reproduce the results of \citet{lample2019cross} and \citet{song2019mass}, we directly run their released codes on the website.\protect\footnotemark[3]}
\end{table*}
\footnotetext[2]{In this paper, we lower-cased all of the case-sensitive languages by default, such as English, German and French.}

\subsection{CSP pre-training}
\textbf{Model configuration}  \quad We choose Transformer as the basic model structure. Following the base model in \citet{vaswani2017attention}, we set the dimension of word embedding as 512, dropout rate as 0.1 and the head number as 8.  To be comparable with previous works, we set the model as 4-layer encoder and 4-layer decoder for unsupervised NMT, and 6-layer encoder and 6-layer decoder for supervised NMT. The encoder and decoder share the same word embeddings. \\
\textbf{Datasets and pre-processing} \quad Following the work of  \citet{song2019mass}, we use the monolingual data sampled from WMT News Crawl datasets for English, German and French, with 50M sentences for each language.\footnotemark[2] For Chinese, we choose 10M sentences from the combination of LDC and WMT2018 corpora. For each translation task, the source and target languages are jointly tokenized into sub-word units with BPE \cite{sennrich-etal-2016-neural}. The vocabulary is extracted from the tokenized corpora and shared by the source and target languages. For English-German and English-French translation tasks, we set the vocabulary size as 32k. For Chinese-English, the vocabulary size is set as 60k since few tokens are shared by Chinese and English. To extract the probabilistic translation lexicons, we utilize the monolingual corpora described above to train the embeddings for each language independently by using word2vec \cite{mikolov2013distributed} . We then apply the public implementation of the method proposed by \citet{Artetxe2017Learning} to map the source and target word embeddings to a shared-latent space.\footnotemark[4] \\
\textbf{Training details} \quad We replace the consecutive tokens in the source input with their translation words sampled from the probabilistic translation lexicons, with random start position $u$. Following \citet{song2019mass}, the length of the replaced fragment is empirically set as roughly 50\% of the total number of tokens in the sentence, and the replaced tokens in the encoder will be the translation tokens 80\% of the time, a random token 10\% of the time and an unchanged token 10\% of the time. \footnotemark[5] In the extracted probabilistic translation lexicons, we only keep top three translation words for each source word and also investigate how the number of translation words produces an effect on the training process.  All of the models are implemented on Py-Torch and trained on 8 P40 GPU cards.\footnotemark[6] We use Adam optimizer with a learning rate of 0.0005 for pre-training. 

\begin{table*}[htb]
			\centering
			  \scalebox{1.0}{
				\begin{tabular}{c|ccc}
					\toprule[2pt]
					 System & en-de	&	en-fr & zh-en\\
					\midrule[1pt]
					\citet{vaswani2017attention} & 27.3& 38.1& -\\
                    \midrule[1pt]
                    \citet{vaswani2017attention} (our reproduction) / + BT & 27.0 / 28.6& 37.9 / 39.3& 42.1 / 43.7\\
                    \citet{lample2019cross} (our reproduction) / + BT & 28.1 / 29.4 & 38.3 / 39.6& 42.0 / 43.7 \\
                    \citet{song2019mass} (our reproduction) / + BT &28.4 / 29.6& 38.4 / 39.6& 42.5 / 44.1  \\
                    \midrule[1pt]
                    \textbf{CSP and fine-tuning (ours)} / + BT & \textbf{28.9 / 30.0}& \textbf{38.8 / 39.9}& \textbf{43.2 / 44.6}  \\
					\bottomrule[2pt]
				\end{tabular}}
				\caption{\label{tab:Result_SNMT} The translation performance of supervised NMT on English-German, English-French and Chinese-to-English test sets. (+ BT: trains the model with back-translation method.)}  
\end{table*}
\footnotetext[3]{\url{https://github.com/facebookresearch/XLM}

\url{https://github.com/microsoft/MASS}}
\footnotetext[4]{The configuration we used to run these open-source tool kits can be found in appendix A.}
\footnotetext[5]{We test different length of the replaced segment and report the results in the appendix B. We find similar results to \citet{song2019mass}.} 
\footnotetext[6]{The code we used can be found in the attached file.} 

\subsection{Fine-tuning on unsupervised NMT}
In this section, we describe the experiments on the unsupervised NMT, where we only utilize monolingual data to fine-tune the NMT model based on the pre-trained model. \\
\textbf{Experimental settings} \quad For the unsupervised English-German and English-French translation tasks, we take the similar experimental settings to \citet{lample2019cross,song2019mass}. Specifically, we randomly sample 5M monolingual sentences from the monolingual data used during pre-training and report BLEU scores on WMT14 English-French and WMT16 English-German. For fine-tuning on the unsupervised Chinese-to-English translation task, we also randomly sample 1.6M monolingual sentences for Chinese and English respectively similar to \citet{yang-etal-2018-unsupervised}. We take $NIST02$ as the development set and report the BLEU score averaged on the test sets $NIST03$, $NIST04$ and $NIST05$. To be consistent with the baseline systems, we apply the script \emph{multi-bleu.pl} to evaluate the translation performance for all of the translation tasks. \\
\textbf{Baseline systems}\quad We take the following four strong baseline systems. \citet{lample-etal-2018-phrase} achieved state-of-the-art (SOTA) translation performance on unsupervised English-German and English-French translation tasks, by utilizing cross-lingual vocabulary, denoising auto-encoding and back-translation. \citet{yang-etal-2018-unsupervised} proposed the weight-sharing architecture for unsupervised NMT and achieved SOTA results on unsupervised Chinese-to-English translation task. \citet{lample2019cross} and \citet{song2019mass} are among the first endeavors to apply pre-training to unsupervised NMT, and both of them achieved substantial improvements compared to the methods without utilizing pre-training. \\
\textbf{Results} \quad Table \ref{tab:Result_UNMT} shows the experimental results on the unsupervised NMT. From Table \ref{tab:Result_UNMT}, we can find that the proposed CSP outperforms all of the previous works on English-to-German, German-to-English, English-to-French and Chinese-to-English unsupervised translation tasks, with as high as +0.7 BLEU points improvement in German-to-English translation task. In French-to-English translation direction, CSP also achieves comparable results with the SOTA baseline of \citet{song2019mass}. In Chinese-to-English translation task, CSP even achieves +1.1 BLEU points improvement compared to the reproduced result of \citet{song2019mass}. These results indicate that fine-tuning unsupervised NMT on the model pre-trained by CSP consistently outperforms the previous unsupervised NMT systems with or without pre-training. 

\begin{figure*}[ht]
   \begin{center}
   \includegraphics[scale=0.35]{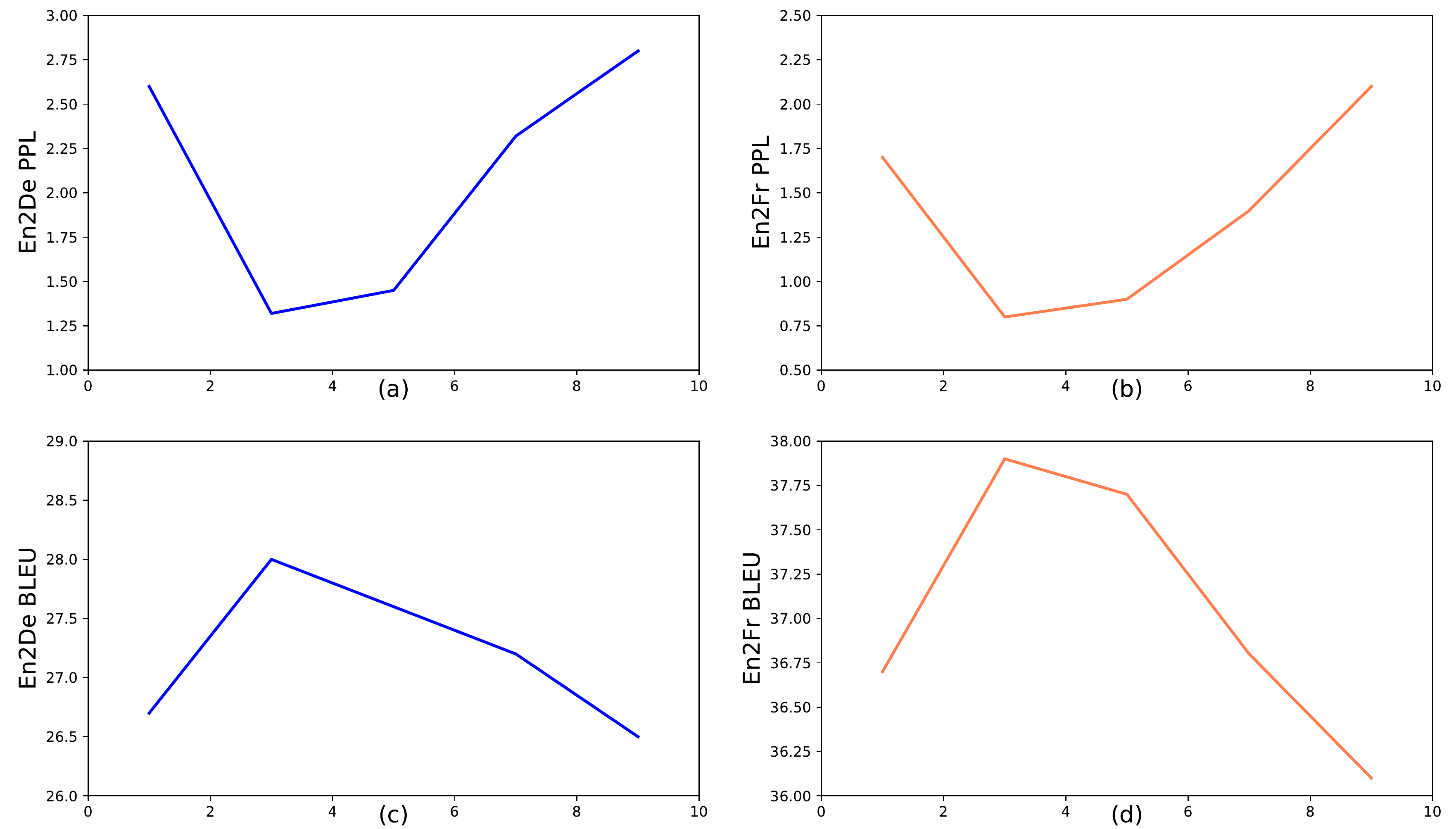}
   \end{center}
   \caption{\label{fig:word_translation} The performance of CSP with the probabilistic translation lexicons keeping different translation words for each source word, which includes: (a) the PPL score of the pre-trained English-to-German model; (b) the PPL score of the pre-trained English-to-French model; (c) the BLEU score of the fine-tuned unsupervised English-to-German NMT model; (d)the BLEU score of the fine-tuned unsupervised English-to-French NMT model.}	
\end{figure*}

\footnotetext[7]{LDC2002L27,LDC2002T01,LDC2002E18,LDC2003E07,
LDC2004T08,LDC2004E12,LDC2005T10}
\footnotetext[8]{Since \textbf{\emph{model-fusion}} approaches incorporate too much extra parameters, it is not fair to take them as baselines here. We leave the comparison between CSP and \textbf{\emph{mode-fusion}} approaches in the appendix C.}
\subsection{Fine-tuning on supervised NMT}
\label{Finetune-super}
This section describes our experiments on supervised NMT where we fine-tune the pre-trained model with bilingual data. \\
\textbf{Experimental settings} \quad For supervised NMT, we conduct experiments on the publicly available data sets, i.e., WMT14 English-French, WMT14 English-German and LDC Chinese-to-English corpora, which are used extensively as benchmarks for NMT systems. We use the full WMT14 English-German and WMT14 English-French corpus as our training sets, which contain 4.5M and 36M sentence pairs respectively. For Chinese-to-English translation task, our training data consists of 1.6M sentence pairs randomly extracted from LDC corpora.\footnotemark[7] All of the sentences are encoded with the same BPE codes utilized in pre-training. \\   
\textbf{Baseline systems} \quad For supervised NMT, we consider the following three baseline systems. \footnotemark[8] The first one is the work of \citet{vaswani2017attention}, which achieves SOTA results on WMT14 English-German and English-French translation tasks. The other two baseline systems are proposed by \citet{lample2019cross} and \citet{song2019mass}, both of which fine-tune the supervised NMT tasks on the pre-trained models. Furthermore, we compare with the back-translation method which has shown its great effectiveness on improving NMT model with monolingual data \cite{sennrich2015improving}. Specifically, for each baseline system, we translate the target monolingual data used 
during pre-training back to the source language by a reversely-trained model, and get the pseudo-parallel corpus by combining the translation with its original data. \footnotemark[9] At last, the training data which includes pseudo and parallel sentence pairs is shuffled and used to train the NMT system. \\
 \textbf{Results} The experimental results on supervised NMT are presented at Table \ref{tab:Result_SNMT}. We report the BLEU scores on English-to-German, English-to-French and Chinese-to-English translation directions. For each translation task, we report the BLEU scores for the standard NMT model and the model trained with back-translation respectively. As shown in Table \ref{tab:Result_SNMT}, compared to the baseline system without pre-training \cite{vaswani2017attention}, the proposed model achieves +1.6 and +0.7 BLEU points improvements on English-to-German and English-to-French translation directions respectively. Even compared to stronger baseline system with pre-training \cite{song2019mass}, we also achieve +0.5 and +0.4 BLEU points improvements respectively on these two translation directions. On Chinese-to-English translation task, the proposed model achieves +0.7 BLEU points improvement compared to the baseline system of \citet{song2019mass}.  With back-translation, the proposed model still outperforms all of the baseline systems. Experimental results above show that fine-tuning the supervised NMT on the model pre-trained by CSP achieves substantial improvements over previous supervised NMT systems with or without pre-training. Additionally, it has been verified that CSP is able to work together with back-translation.

\footnotetext[9]{We randomly select the target monolingual data with the same size to the bilingual data.}

\section{Analysis}
\subsection{Study the number of translation words}
In CSP, the probabilistic translation lexicons only keep the top $k$ translation words for each source word. For each word in the translation lexicons, the number of translation words $k$ is viewed as an important  hyper-parameter and can be set carefully during the process of pre-training. A natural question is that how much of translation words do we need to keep for each source word? Intuitively, if $k$ is set as a small number, the model may lose its generality since each source word can be replaced with only a few translation words, which severely limits the diversity of the context. And if otherwise, the accuracy of the extracted probabilistic translation lexicons may get significantly diminished, which shall introduce too much noise for pre-training. Therefore, there is a trade-off between the generality and accuracy. We investigate this problem by studying the translation performance of unsupervised NMT with different $k$, where we vary $k$ from 1 to 10 with the interval 2.  We observe both the performance of CSP after pre-training and the translation performance after fine-tuning on the unsupervised NMT tasks, including the English-to-German and English-to-French translation directions. For each translation direction, we firstly present the perplexity (PPL) score of the pre-trained model averaged on the monolingual validation sets of the source and target languages.\footnotemark[10] And then we show the BLEU score of the fine-tuned model on the bilingual validation set.  Figure \ref{fig:word_translation} (a) and (c) illustrate the PPL score of the pre-trained model and BLEU score of the fine-tuned unsupervised NMT model respectively on English-to-German translation. Figure \ref{fig:word_translation} (b) and (d) present the PPL and BLEU score respectively for English-to-French translation. From Figure \ref{fig:word_translation}, it can be seen that, when $k$ is set around 3, the pre-trained model achieves the best validation PPL scores on both of the English-to-German and English-to-French translation directions. Similarly, CSP also achieves the best BLEU scores on the unsupervised translation tasks when $k$ is set around 3.
\footnotetext[10]{For English-German translation, the monolingual validation set for English is built by including all English sentences in the bilingual English-German validation set, and the monolingual validation set for German is built in the same way. }
\subsection{Ablation study}
To understand the importance of different components of the model pre-trained by CSP, we perform an ablation study by training multiple versions of the supervised NMT model with some components initialized randomly: the word embeddings, the encoder, the attention module between the encoder and decoder, and the decoder. Experiments are conducted on English-to-German and English-to-French translation tasks. All models are trained without back-translation and results are reported in Table \ref{tab:ablation-study}. We can find that the two most critical components are the pre-trained encoder and attention module. It shows that CSP enhances NMT not only on the ability of building sentence representation for the input sentence, but also on the ability of aligning the source and target languages with the help of word-pair alignment information. Additionally, the experimental results indicate that the pre-trained decoder shows little effect on the translation performance. This is mainly because the decoder only predicts the source-side words during pre-training but predicts the target-side words during fine-tuning. This pretrain-finetune mismatch makes the pre-trained decoder less helpful for performance improvement. 
\begin{table}[htb]
			\centering
			\scalebox{0.93}{
				\begin{tabular}{c|cc}
					\toprule[2pt]
					 System & en-de	&	en-fr \\
					\midrule[1pt]
					No pre-trained embeddings & 28.4 & 38.5 \\
					No pre-trained encoder &27.9 & 38.2\\
                    No pre-trained attention module &28.1 &38.3 \\
                    No pre-trained decoder & 28.8& 38.8\\
                    Full model pre-trained by CSP & 28.9 & 38.8  \\
					\bottomrule[2pt]
				\end{tabular}}
				\caption{\label{tab:ablation-study} Ablation study on English-German and English-French translation tasks. The embeddings include the source-side and target-side word embeddings.}  
\end{table}

\subsection{Code-switching translation}
Code-switching, which contains words from different languages in single input, has aroused more and more attention in NMT \cite{kwak2016google,menacer2019machine}.
In this section, we show that the proposed CSP is able to enhance the ability of the fine-tuned NMT model on handling the code-switching input. To present quantitative results, we build two test sets for the supervised Chinese-to-English translation task to evaluate the performance of the translation model on handling code-switching inputs. We randomly select 200 Chinese-English sentence pairs from $NIST02$, based on which we build two code-switching test sets. The first test set, referred to as test A, is built by randomly replacing some phrases in each Chinese sentence with their counterpart English phrases, where the English phrase is the translation result by feeding the corresponding Chinese phrase to the Google Chinese-to-English translator; The second test set, referred to as test B, is constructed by randomly replacing parts of the words in each Chinese sentence with their nearest target words in the shared latent embedding space (the same way used by CSP in Section \ref{ptl}).  Table \ref{tab:code-switch-result} shows the translation performance of NMT systems on the two code-switching test sets.\footnotemark[11] Besides the baseline systems mentioned in section \ref{Finetune-super}, we also train a Chinese-English multi-lingual system \cite{kwak2016google} based on Transformer, which has shown the ability of handling code-switching inputs. From Table \ref{tab:code-switch-result}, We can find that the proposed approach achieves significant improvements over previous works. Compared to multi-lingual system, we achieve +2.3 and +3.0 BLEU points improvements respectively on test A and test B. The case study can be found in appendix D.
\begin{table}[htb]
			\centering
			\scalebox{0.93}{
				\begin{tabular}{c|cc}
					\toprule[2pt]
					 System & test A	&	test B\\
					\midrule[1pt]
					\citet{vaswani2017attention} & 28.17& 32.51\\
                    \citet{lample2019cross}& 28.82 & 32.90 \\
                    \citet{song2019mass}&28.70 & 33.21\\
                    Multi-lingual system & 30.51& 35.10 \\
                    \midrule[1pt]
                    \textbf{CSP and fine-tuning} & \textbf{32.84}& \textbf{38.17}\\
					\bottomrule[2pt]
				\end{tabular}}
				\caption{\label{tab:code-switch-result} The performance of Chinese-to-English translation on in-house code-switching test sets.}  
\end{table}

\footnotetext[11]{The two in-house code-switching test sets can be found in the attached files.}

\section{Conclusions and Future work}
This work proposes a simple yet effective pre-training approach, i.e., CSP for NMT, which randomly replaces some words in the source sentence with their translation words in the probabilistic translation lexicons extracted from monolingual corpus only. To verify the effectiveness of CSP, we investigate two downstream tasks, supervised and unsupervised NMT, on English-German, English-French and Chinese-to-English translation tasks. Experimental results show that the proposed approach achieves substantial improvements over strong baselines consistently. Additionally, we show that CSP is able to enhance the ability of NMT on handling code-switching inputs. There are two promising directions for the future work. Firstly, we are interested in applying CSP to other related NLP areas for code-switching problems. Secondly, we plan to investigate the pre-training objectives which are more effective in utilizing the cross-lingual alignment information for NMT.

\section{Acknowledgement}
We sincerely thank the anonymous reviewers for their thorough reviewing and valuable suggestions.

\bibliographystyle{acl_natbib}
\bibliography{anthology,emnlp2020}

\end{document}